\pdfoutput=1
\documentclass[runningheads]{llncs}
\usepackage{graphicx}
\usepackage{amsmath,amssymb} 
\usepackage{color}
\usepackage[width=122mm,left=12mm,paperwidth=146mm,height=193mm,top=12mm,paperheight=217mm]{geometry}
\usepackage[pagebackref=true,breaklinks=true,letterpaper=true,colorlinks,bookmarks=false]{hyperref}
\usepackage{multirow}
\usepackage{wrapfig}
\usepackage[tight,footnotesize]{subfigure}
\newcommand{\etal}{\textit{et al}. }
\begin{document}
\pagestyle{headings}
\mainmatter

\title{Video Frame Interpolation by Plug-and-Play Deep Locally Linear Embedding} 

\author{Anh-Duc Nguyen, Woojae Kim, Jongyoo Kim, and Sanghoon Lee}
\institute{Yonsei University}

\maketitle

\begin{abstract}
We propose a generative framework which takes on the video frame interpolation problem. 
Our framework, which we call Deep Locally Linear Embedding (DeepLLE), is powered by a deep convolutional neural network (CNN) while it can be used instantly like conventional models. DeepLLE fits an auto-encoding CNN to a set of several consecutive frames and embeds a linearity constraint on the latent codes so that new frames can be generated by interpolating new latent codes. Different from the current deep learning paradigm which requires training on large datasets, DeepLLE works in a plug-and-play and unsupervised manner, and is able to generate an arbitrary number of frames. Thorough experiments demonstrate that without bells and whistles, our method is highly competitive among current state-of-the-art models.
\end{abstract}

\section{Introduction}


Video frame interpolation is among the most long-standing and challenging problems in computer vision. Traditionally, optical flow based and phase based methods are extensively studied to deal with this problem. However, while optical flow estimation is another classic and difficult task per se, phase based methods are proven to be suitable for small motion videos only. Nonetheless, a merit of these learning-free methods is that they are off-the-shelf models and can be used instantly, and hence they always work at their full capacities. 

Recent years have seen a huge success of convolutional neural networks (CNNs), especially in the large-scale ImageNet classification challenge \cite{rn142}. Since then, researchers have considerably brought them into use in many different computer vision tasks including frame synthesis. The performances of deep learning based methods are superior to those of conventional algorithms thanks to their phenomenal generalization abilities, but unfortunately, training on a large-scale dataset is usually required beforehand, which is generally time-consuming and infeasible in many situations. Also, to make them work at their fullest potentials, adaptation or fine-tuning might be required since in training, one may impose several regularizations on the models, which makes them biased estimators.

\begin{wrapfigure}{R}{0.5\textwidth}
\centering
\includegraphics[width=0.5\textwidth]{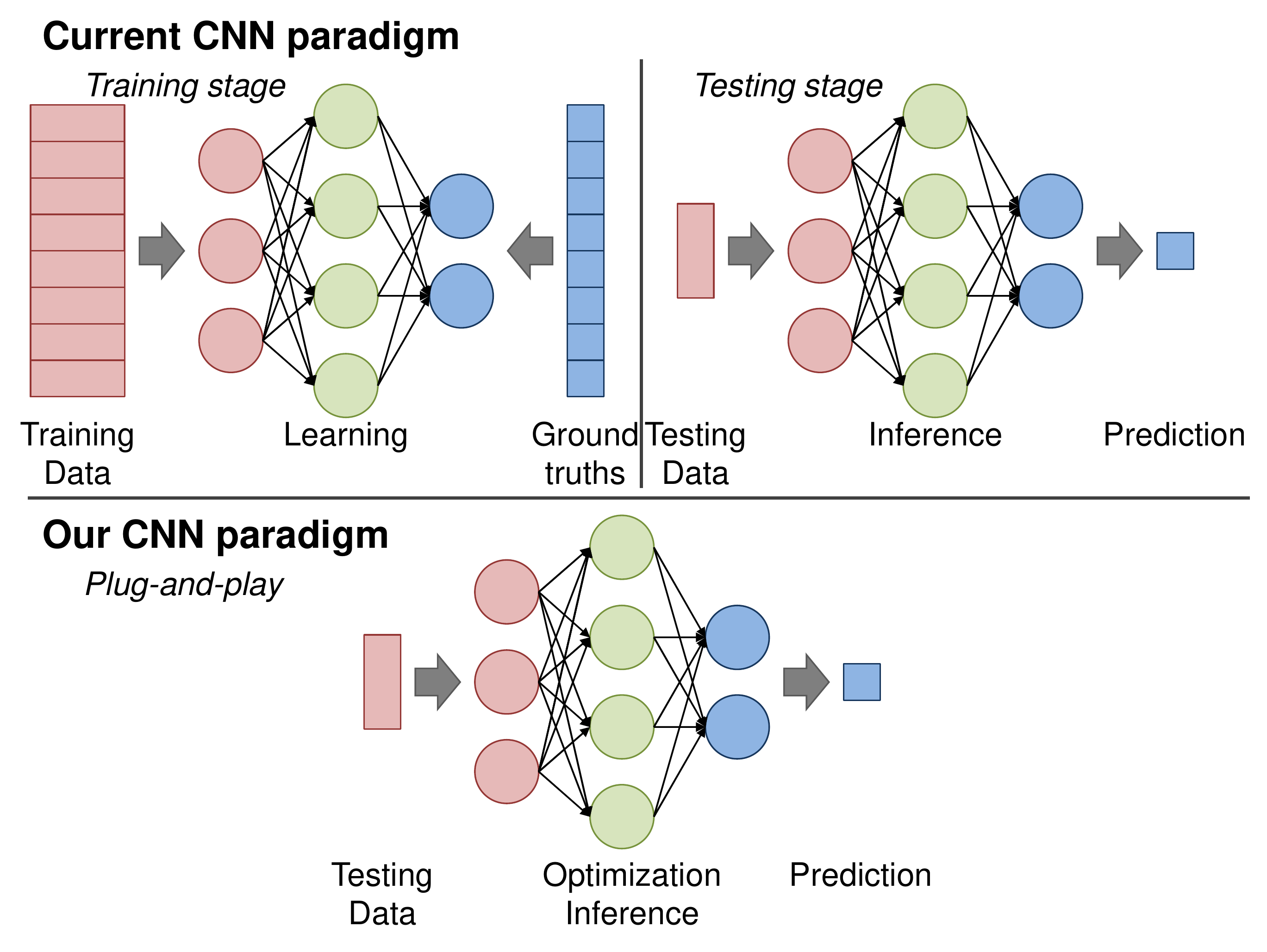}
\caption{The difference between an usual CNN paradigm and ours.}
\label{paradigm}
\end{wrapfigure}

In this paper, we propose a novel framework of video frame interpolation, dubbed as Deep Locally Linear Embedding (DeepLLE), which combines the advantages of conventional models and deep networks to address the frame synthesis problem. Different from previous deep learning based studies in frame synthesis, our method works in a plug-and-play and unsupervised fashion like conventional methods. DeepLLE is built upon a hypothesis that in some latent space, consecutive frames lie very close to each other on a manifold and we can explicitly embed a linearity constraint to the latent codes so that new frames can be produced by interpolating new latent codes. To do so, we resort to the computing power of an auto-encoding CNN. 

Figure~\ref{paradigm} illustrates the difference between our scheme and existing models. While current trends require a separate training stage for CNNs, our method can perform both optimization and inference on the fly. Compared with existing methods, ours has a number of advantages. First, DeepLLE is an instant method like conventional ones while having a performance level of deep learning. Second, DeepLLE can synthesize new frames between any number of successive frames greater than one. Lastly, DeepLLE can generate an arbitrary number of frames in a run. To the best of our knowledge, no existing method can include all these properties at the same time. Our network is also optimized with a perceptual cost function so the synthesized images have a much better perceptual quality than those produced by existing methods.

\textbf{Contributions.} Our work mainly focuses on the following contributions. First, we propose a new frame synthesis framework that bridges the gap between conventional methods and deep learning based models. The method is plug-and-play, which always allows it to work at its full potential, but still possesses a deep-learning-standard performance.
Next, different from existing methods, ours is entirely based on manipulating the underlying latent structures of videos, which has not been successfully applied to real-world videos consisting of complex motions. Finally, we discover that deep networks can capture a great deal of video statistics, which may not only be beneficial to the frame synthesis field but also advance the deep unsupervised learning area.

\section{Related Work}
Video frame interpolation is a classic problem in video processing. Traditionally, this is done by estimating motions in consecutive frames \cite{rn146,rn147}. Mahajan \etal defined ``path", which is optical flow-like, and then used 3D Poisson reconstruction for in-between frame generation \cite{rn148}. Alternative to optical flow based methods, Meyer \etal utilized phase information to interpolate frames but this method may fail to retain high frequency details in regions containing large motions \cite{rn139}.

Since the huge success of deep CNNs in image recognition/classification \cite{rn25,rn153,rn64}, many researchers have proposed to estimate optical flow using CNNs \cite{rn149,rn150,rn151}. These methods are not optimized to directly synthesize new images, and so the generated images may contain many artifacts due to inaccurate flows and warping. The recent state-of-the-art Deep Voxel Flow (DVF) \cite{rn132} unsupervisedly estimates a voxel flow field, which is optical flow from the interpolated image to the previous and next frames, and trilinearly interpolates pixels in the new frame. In a different direction, the method in \cite{rn143} is based on a technique called pixel hallucination which directly generates new pixels from scratch. However, perceptually, the results are not good-looking \cite{rn132}. Lastly, a recent technique based on adaptive convolution was proposed in \cite{rn134,rn161}. These methods learn to produce a set of pixel-dependent filters and convolve these filters with the input frames to interpolate each pixel. The results are notable but the method requires to learn a set of filters for each pixel, which is computationally unfriendly. Above all, all these methods require a separate training step on big data and might need fine-tuning or adaptation to take full advantage of the networks.

The technical inspiration for our work is Roweis and Saul \cite{rn138}. The method finds a linear relationship between neighboring samples in their original space and then finds another space, preferably having lower dimension, where this relationship is preserved. We, however, do not impose any linearity on images but only on their latent codes and this linear relationship is explicitly specified instead of learning from data as in \cite{rn138}. Another interpolation research close to ours is proposed by Bregler \etal which also relies on manifold learning \cite{rn144}. However, they estimate a non-linear manifold for the whole series of frames and the images are simply talking lips. Here, the method is designed to work with natural images containing different kinds of motion. In addition, our method does not estimate the manifold of the whole video but several consecutive frames only. In a remote study \cite{rn137}, an observation that latent variables can encode motions backs up our idea, but their latent codes are randomly sampled from a Gaussian distribution, and their work concentrates on predicting the trajectory of motions given a still image, and hence the generated images are nowhere near realistic.

\section{Locally Linear Embedding}
To make the paper self-contained, we briefly summarize the idea of locally linear embedding (LLE) \cite{rn138}, and then we highlight key concepts that inspire our approach. Suppose we have a dataset $X \in {\mathbb{R}^{m \times n}}$ having $m$ samples of $n$ dimensions. We want to find a compact representation of $X$, \emph{i.e.}, a lower-dimensional space that preserves local relationships of each point. We assume that there are sufficient data so that all the twists and variations in the manifold are accounted for. Having this condition, we hypothesize that each point is a linear combination of those in its neighborhood. To find such a linear relationship $W$, for each $X_i$, we find a set $N_i$ containing all the indices of the neighbors of $X_i$ according to some distance threshold and then solve the following optimization to find the weights $W_{ij}$ 

\begin{equation} \label{lle1}
\begin{gathered}
  \mathop {\min }\limits_W {\sum\limits_i {\left\| {{X_i} - \sum\nolimits_{j \in {N_i}} {{W_{ij}}{X_j}} } \right\|} ^2}, \hfill \,
  s.t\,\,\,\sum\limits_j {{W_{ij}}}  = 1 \hfill
\end{gathered} ,
\end{equation} 
where $W_{ij}=0$ if $X_j \notin N_i$. The goal of LLE is to find a new low-dimensional space in which the linear relationship $W$ is preserved. Suppose we have a function which maps each $X_i$ in the original space to each $Y \in {\mathbb{R}^{m \times k}}$ in some latent space where $k \ll n$, then we can find $Y$ by solving the following problem

\begin{equation} \label{lle2}
\mathop {\min }\limits_Y \sum\limits_i {{{\left\| {{Y_i} - \sum\nolimits_{j \in {N_i}} {{W_{ij}}{Y_j}} } \right\|}^2}} ,
\end{equation}
which is similar to (\ref{lle1}) but the minimization is over $Y$ instead of $W$.


Applying LLE directly to frame interpolation, however, is infeasible due to two reasons. First, like other manifold learning methods, LLE requires abundant data so that the manifold is well-sampled. Unlike facial expression images in \cite{rn138} or talking lips in \cite{rn144}, real-world videos do not contain any common objects or structures. Given a target frame, only a few neighboring frames are useful for estimating manifold. Second, in order to synthesize new images, we can somehow synthesize some new $Y_k$ and $W_{kj}$, and use this $W_{kj}$ on $X$ to synthesize a new image. However, finding $W_{kj}$ may involve an optimization process, and it is not clear how we can synthesize $Y_k$ corresponding to a desired $X_k$ in the first place. Nevertheless, LLE gives us several key ideas that motivate our framework. First, we do not need to work with the whole video but several successive frames only, and assume their underlying manifold in some latent space is linear. Second, if we can encode video frames into this latent space and then map their codes back into image space, then synthesizing new frames can simply be done in the latent space. In Section~\ref{DeepLLE}, we depict how DeepLLE is built based on this idea.

\section{Deep Locally Linear Embedding} \label{DeepLLE}

\begin{figure}[tp]
\centering
\includegraphics[width=0.65\textwidth]{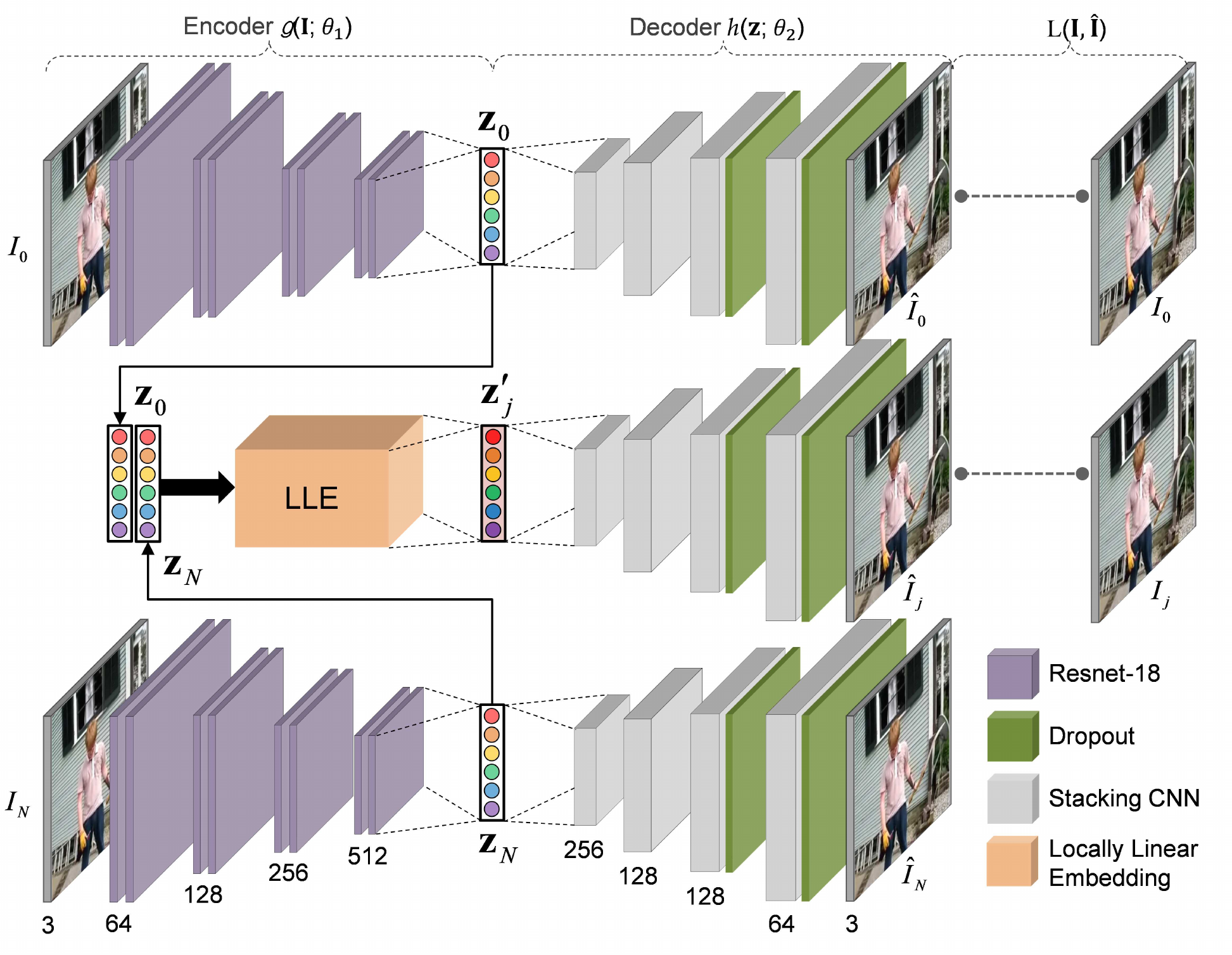}
\caption{Overall framework of DeepLLE. First, the encoder $g$ maps two nodes into a latent space where they are fed to an LLE module to interpolate a new latent code. Then, all the latent codes are decoded by the decoder $h$ to reconstruct the reference images. The three streams of network share weights with each other.}
\label{framework}
\end{figure}

\subsection{Our framework} \label{framework section}
\begin{wrapfigure}{R}{0.5\textwidth}
\centering
\includegraphics[width=0.5\textwidth]{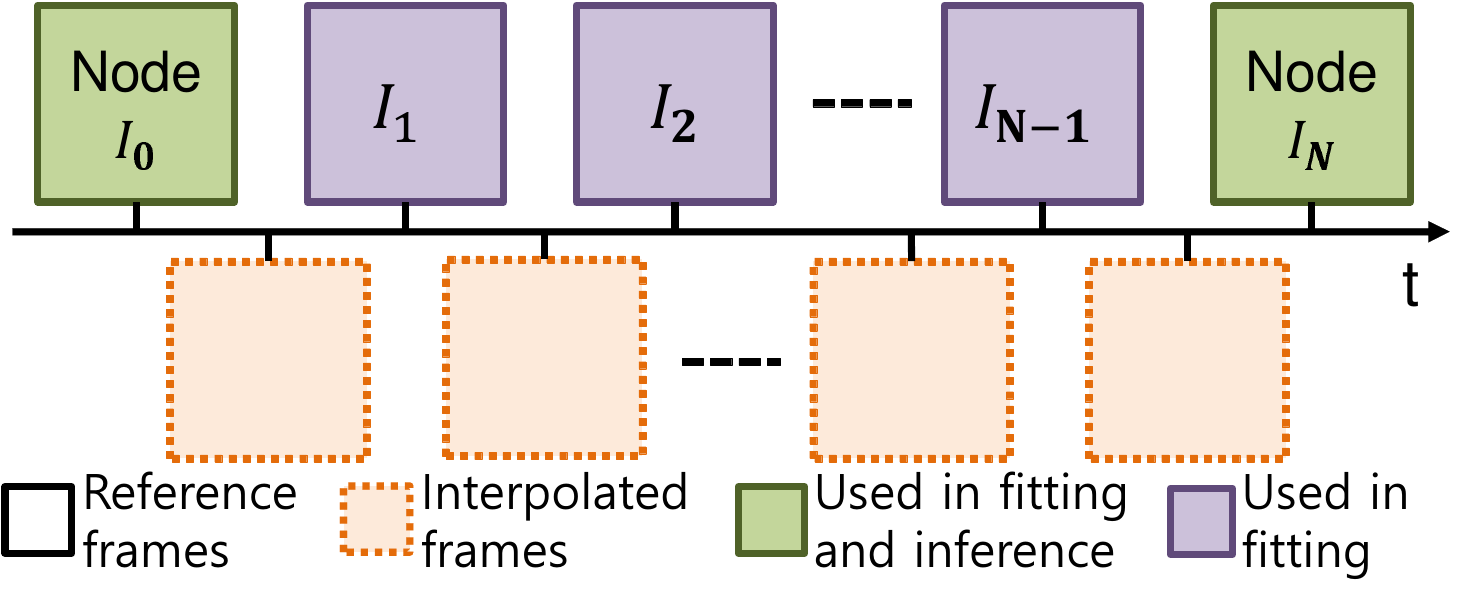}
\caption{Notations of video frames.}
\label{input}
\end{wrapfigure}

The overall framework of DeepLLE is shown in Figure~\ref{framework}. Suppose we have $N+1$ consecutive frames $I_0,I_1,...,I_N$ which are temporally uniform as in Figure~\ref{input}. We consider $N>1$ for now. We refer to these frames as \textit{references}, and $I_0$ and $I_N$ as \textit{nodes}. Different from previous studies, we do not assume that motions are symmetric over the middle frame or all frames are stabilized with respect to the starting frame. Our method first performs a fitting process to construct a linear manifold from the given nodes so that certain points on this manifold can reconstruct the references. After that, the constructed manifold is used to interpolate new frames between two nodes. 

\begin{wrapfigure}{R}{0.5\textwidth}
	\centering
	\includegraphics[width=0.5\textwidth]{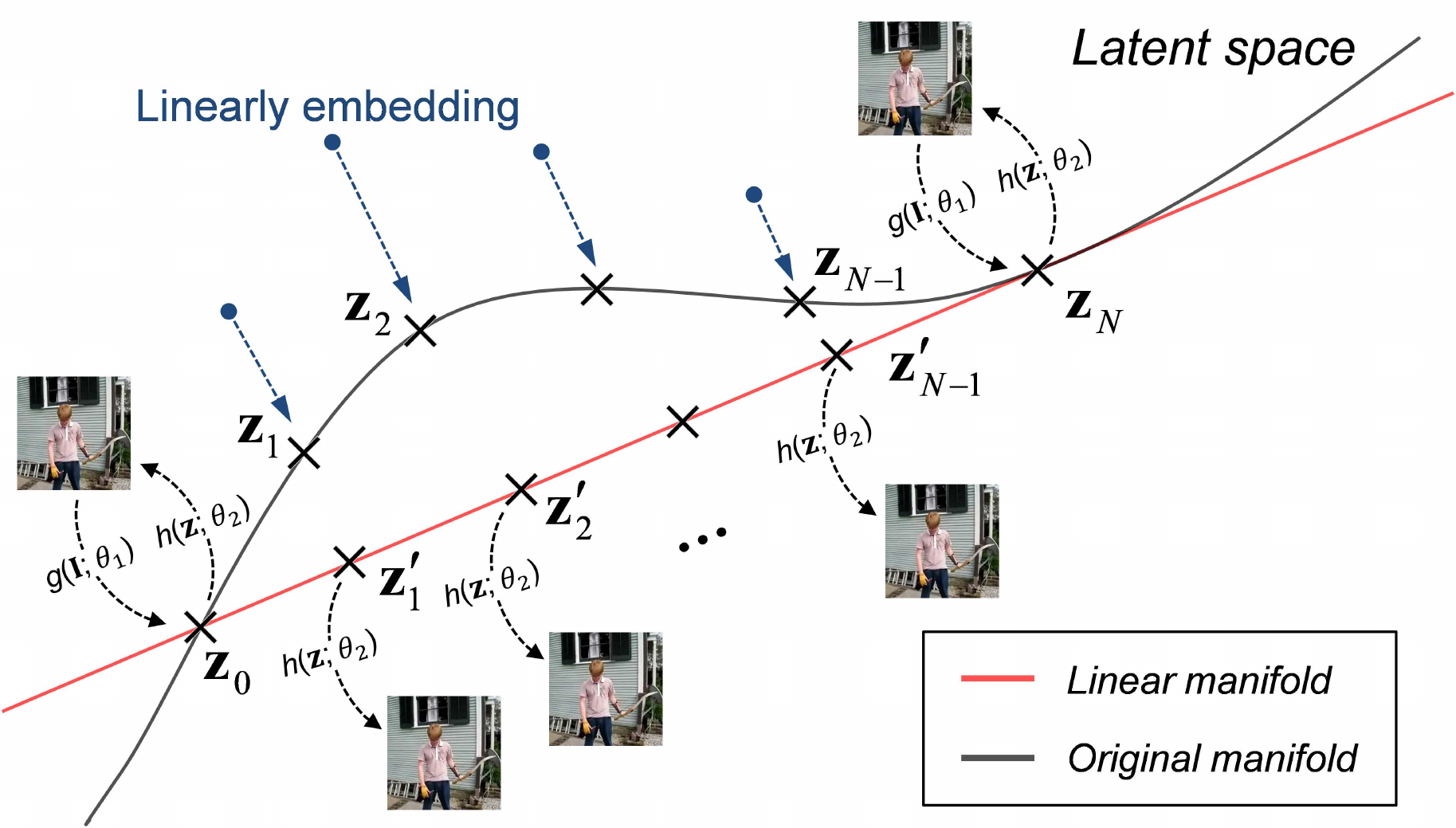}
	\caption{Linear embedding in latent space. Latent codes are linearly interpolated and evenly distributed on the line segment between the codes of the two nodes.}
	\label{latent}
\end{wrapfigure}

We now describe the fitting process. As can be seen from Figure~\ref{framework}, at a first glance, DeepLLE simply fits a CNN to a set of input images. The encoder $g$ of DeepLLE first maps $I_i$ into $z_i \in {\mathbb{R}^{k}}$ in some $k$-dimensional latent space for $i \in \{ 0,N\}$. Then, the decoder $h$ decodes $z_0$ and $z_N$ into $\hat I_0$ and $\hat I_N$, respectively. The crucial ingredient added to the fitting procedure is an LLE module in the decoding process of the frames in-between $I_0$ and $I_N$. Concretely, for some $0<j<N$, instead of calculating and decoding its own latent code, the decoder decodes $z'_j$, which is the output of the LLE module, to reconstruct $\hat I_j$. This module is similar to LLE; \emph{i.e.}, it linearly combines $z_0$ and $z_N$ to produce $z'_j$. However, in LLE, the weights are determined through an optimization process. Such complicated weights are not suitable in our method since LLE is designed to reduce data dimension, not to interpolate new latent variables and produce new data. Instead, we manually define the weights in a more intuitive way. An illustration of our strategy is shown in Figure~\ref{latent}. We define the weights based on the relative temporal position of $I_j$ with respect to the two nodes so that all $z_j$ are uniformly distributed between $z_0$ and $z_N$ and in the same position as the temporal position of frame $I_j$ with respect to the two nodes. Specifically, $z'_j$ is calculated as

\begin{equation} \label{latent interp}
{{z'}_j} = z_0 + \frac{j}{N}{(z_N - z_0)} ,
\end{equation}
that is, the new latent codes lie evenly on the line segment between $z_0$ and $z_N$. Obviously, this is an oversimplification of the underlying manifold. The real manifold in any non-trivial case should be highly non-linear and there is no guarantee whatsoever that the latent codes of the references are co-planar, much less on the same line. However, this is a reasonable choice because it is difficult to walk along the manifold if the manifold is complicated, and this oversimplification is very intuitive when it comes to choosing a frame position to interpolate.

We can generalize the framework into a matrix form. Let ${\mathbf{I}} \in {\mathbb{R}^{(N+1) \times c \times h \times w}}$ be a 4-D tensor containing $N$ consecutive frames having height $h$, width $w$, and $c$ channels, and ${\mathbf{I_{nodes}}} \in {\mathbb{R}^{2 \times c \times h \times w}}$ contains the first and last frames of the sequence. First, the encoder maps the two nodes to a latent space in which the images are 

\begin{equation}
Z_{nodes} =g(\mathbf{I_{nodes}};\theta_1),
\end{equation}
where ${Z_{nodes}} = {\left( {\begin{array}{*{20}{c}}
  {{z_0}}&{{z_N}} 
\end{array}} \right)^T} \in {\mathbb{R}^{2 \times k}}$ contains the latent code of dimension $k$ of each node in its row, and $\theta_1$ are trainable parameters of the encoder. Next, we perform linear interpolation on the latent codes based on the relative temporal position of the reference frames with respect to the nodes. Towards this goal, we define a relative position matrix (RPM) $M \in {\mathbb{R}^{N \times 2}}$ which interpolates new latent variables from the two nodes' codes. To fit the network to the reference frames, $M$ is defined as

\begin{equation} \label{relative position matrix}
M = {\left( {\begin{array}{*{20}{c}}
  1&{\frac{{N - 1}}{N}}& \cdots &{\frac{{N - j}}{N}}& \cdots &{\frac{1}{N}}&0 \\ 
  0&{\frac{1}{N}}& \cdots &{\frac{j}{N}}& \cdots &{\frac{{N - 1}}{N}}&1 
\end{array}} \right)^T} ,
\end{equation}
where $0<j<N$, and $T$ denotes the transpose of a matrix. The coefficients are determined based on the relative temporal position of each reference frame with respect to the nodes. Also, the sum of each row is equal to $1$, which has the same spirit as LLE. The interpolation can then be expressed as 

\begin{equation}
Z' = MZ_{nodes}
\end{equation}
where $Z'$ contains the interpolated codes of the reference frames in its rows. Finally, the interpolated codes are decoded back into the image space to become

\begin{equation}
{{\mathbf{\hat I}}} = h(Z';\theta_2),
\end{equation}
where ${{\mathbf{\hat I}}}$ is the reconstructed versions of the reference frames, and $\theta_2$ parametrize the decoder of DeepLLE. Like a common auto-encoder framework, we minimize some expected loss between $\mathbf{\hat I}$ and $\mathbf{I}$ for optimal $\theta_1$ and $\theta_2$. In practice, the fitting is carried out blindly without any stopping criterion based on a development set so the optimization is very close to function approximation in its true sense.

After the fitting is done, to generate new frames in-between $I_j$ and $I_{j+1}$ for any $j$, we can simply define a suitable RPM. For example, suppose we desire to synthesize a new frame halfway in temporal order between $I_j$ and $I_{j+1}$, then we can simply define a new RPM $M'$ and the new latent code can be calculated as 

\begin{equation}
z' = z_j + (z_{j+1} - z_j)/2 = M'Z,
\end{equation}
where $M' = \left( {\begin{array}{*{20}{c}}
  {\frac{{2N - 2j - 1}}{{2(N - 1)}}}&{\frac{{2j + 1}}{{2(N - 1)}}} 
\end{array}} \right)$. $M'$ intuitively reorders all the latent codes according to the relative positions of the input frames with respect to the nodes. Therefore, we can interpolate any frame in-between the two nodes by simply reflecting its relative position with respect to the two nodes to the RPM. Needless to say, to synthesize any number of new frames, we can horizontally stack all RPMs into a big RPM. Moreover, the RPM can be tweaked so that fitting can be performed for more than one sequence of reference frames at a time\footnotemark.

Note that the RPM has nothing to do with optical flow. Different from optical flow-based methods, our approach does not have any control over how many pixels an object should move. Setting the RPM so that the latent code is midway between two reference codes does not guarantee the generated motion is symmetric around the interpolated frame, but it does make sure that the synthesized frame is halfway between the two reference frames in temporal order. The RPM is simply a convenient and intuitive way that we embed a linearity constraint on the underlying manifold instead of learning a linear relationship in the neighborhood of each point like in LLE. 

\footnotetext{Please check the supplementary materials. \label{footnote supp}}

\subsection{Architecture} \label{learning}
Our model works in a plug-and-play and unsupervised manner so the only required input is a set of several consecutive frames. As introduced before, DeepLLE employs an auto-encoding CNN to go back and forth between image space and latent space. We use the 18-layer deep residual network (ResNet-18) \cite{rn153} as the encoder of DeepLLE after removing the first mean pooling layer, the global average pooling layer and the softmax layer. Another modification is that we replace the rectified linear unit (ReLU) \cite{rn152} activation by leaky ReLU (LReLU) \cite{rn29} with the leaky parameter of $0.1$. The reason is that we do not want the network to be robust against small changes in its input, which is brought by functions having saturating regions \cite{rn156}. Another reason is that it is more linear than all of its siblings\textsuperscript{\ref{footnote supp}}. We perform extensive experiments to verify this choice. For the decoder part, we found that simply stacking up convolutional layers interleaved with bicubic upsampling works consistently well in all our simulations. The detail of the decoder is described in Figure~\ref{framework}. The numbers of convolutional layers in each stacking block are 3, 5, 7 and 9, respectively. All the kernels in the decoder have a receptive field size of $5 \times 5$. LReLU is used in all layers except for the output layer which is activated by the hyperbolic tangent function. The reconstructed images are simply scaled by ${\mathbf{\hat I}} = {{\mathbf{Y}} \mathord{\left/
 {\vphantom {{\mathbf{Y}} 2}} \right.
 \kern-\nulldelimiterspace} 2} + 0.5$, where $\mathbf{Y}$ is the output tensor from the decoder. Interestingly, we discovered that dropout \cite{rn67} reduces glaring artifacts of the generated images very efficiently. Therefore, we add two dropout layers with a dropout probability of $0.5$ after the third and fourth stacking modules. We note that a more sophisticated choice of the network structure may improve performance but is not the main issue of our work.

To optimize DeepLLE, we use Huber loss with a threshold of $0.01$ on the reconstructed images and reference frames. Together with Huber loss, we optimize the network with a first derivative loss and the structural similarity score (SSIM) \cite{rn154} between the output images and ground truths. For SSIM, we first convert the output images and ground truths to YCbCr color space and apply SSIM to the first channels only. Overall, our objective function is 

\begin{equation}
\operatorname{L} = \operatorname{H} ({\mathbf{I}},{\mathbf{\hat I}}) + {\lambda _1}{\left\| {\nabla {\mathbf{I}} - \nabla {\mathbf{\hat I}}} \right\|_1} - {\lambda _2}\operatorname{SSIM} ({\text{Y}_{\mathbf{I}}},{{\text{Y}}_{{\mathbf{\hat I}}}}) ,
\end{equation}
where $\text{Y}_{\mathbf{I}}$ and ${\text{Y}}_{{\mathbf{\hat I}}}$ are the Y channel of $\mathbf{I}$ and $\mathbf{\hat I}$, respectively; $\nabla$ denotes the gradient operator, $\operatorname{H}(\cdot,\cdot)$ stands for the Huber loss, $\operatorname{SSIM} ( \cdot , \cdot )$ indicates the SSIM metric, and ${\left\|  \cdot  \right\|_p}$ is the $L_p$-norm. We empirically set $\lambda_1=0.1$ and $\lambda_2=0.0001$ in all experiments. The network is optimized end-to-end so that the decoder can guide the encoder to find a latent space where both the reconstructions from the nodes' latent codes and the interpolated codes are possible. We initialize the weights of the auto-encoder using He initialization \cite{rn155}. We use the ADAM optimization scheme \cite{rn65} with a learning rate of $10^{-4}$ and other parameters are set to the authors' suggestions. Since the fitting is blindly operated, we can either run until convergence or terminate the optimization at some fixed iteration. In our experiments, we chose the latter and set the number of iterations to $5000$. As a side note, better performance might be achieved by running the optimization longer. We implemented our model using Theano\footnote{The code will be released upon the publication of the paper.} \cite{rn1}.

\section{Experimental Results}
We assessed our framework using the UCF-101 \cite{rn135}, DAVIS \cite{rn136,rn160} and real-life videos. For UCF-101, we tested our method on the test set provided in \cite{rn143}. In each sequence, there are $7$ consecutive frames and we took the last $5$ frames as references. To quantitatively evaluate our method, for each sequence, we left out all the even frames to compare the interpolated frames with. We emphasize that the quantitative evaluation is unfair to our method due to two reasons. First, we have to skip all even frames as ground truths for evaluation, and so in the optimization process, the frames are farther from each other compared with the inputs of other methods, which hurts our linearity assumption. Second, our method is not directly trained to interpolate a frame halfway between two inputs but to generate any frame between two nodes, and so the generated motions are likely to be different from those in ground truths. Therefore, the quantitative results do not reflect the full potential of our method. Nonetheless, we used five metrics, namely visual information fidelity (VIF) \cite{rn162}, detail loss metric (DLM) \cite{rn165}, noise quality measure (NQM) \cite{rn164}, SSIM and PSNR, which are widely used in image quality assessment literature, to evaluate the synthesized images\footnote{Evaluations of DVF except for SSIM and PSNR were done on their provided results.}. Following \cite{rn132,rn143}, we evaluated on the motion regions only, which are extracted by applying the masks in \cite{rn143}. All input images were processed at their original resolution ($240\times 320$) and normalized into the range of $[-1,1]$. We chose 
$
M' = \left( {\begin{array}{*{20}{c}}
  {{\raise0.7ex\hbox{$3$} \!\mathord{\left/
 {\vphantom {3 4}}\right.\kern-\nulldelimiterspace}
\!\lower0.7ex\hbox{$4$}}}&{{\raise0.7ex\hbox{$1$} \!\mathord{\left/
 {\vphantom {1 4}}\right.\kern-\nulldelimiterspace}
\!\lower0.7ex\hbox{$4$}}} \\ 
  {{\raise0.7ex\hbox{$1$} \!\mathord{\left/
 {\vphantom {1 4}}\right.\kern-\nulldelimiterspace}
\!\lower0.7ex\hbox{$4$}}}&{{\raise0.7ex\hbox{$3$} \!\mathord{\left/
 {\vphantom {3 4}}\right.\kern-\nulldelimiterspace}
\!\lower0.7ex\hbox{$4$}}} 
\end{array}} \right)$ to interpolate a frame half-way between each two reference frames. For DAVIS, we selected several videos and only measured the performance visually since there is no previous work performing any benchmark on this database. To show off the ability to generate any number of frames in-between two nodes, we interpolated 3 new frames between each two reference frames. We applied the same strategy to several real-life video segments including discoveries, music videos and movies. Due to the limited GPU memory, all sequences are processed at $224\times 384$. We chose $M' = {\left( {\begin{array}{*{20}{c}}
  {{\raise0.7ex\hbox{$5$} \!\mathord{\left/
 {\vphantom {5 6}}\right.\kern-\nulldelimiterspace}
\!\lower0.7ex\hbox{$6$}}}&{{\raise0.7ex\hbox{$3$} \!\mathord{\left/
 {\vphantom {3 4}}\right.\kern-\nulldelimiterspace}
\!\lower0.7ex\hbox{$4$}}}&{{\raise0.7ex\hbox{$2$} \!\mathord{\left/
 {\vphantom {2 3}}\right.\kern-\nulldelimiterspace}
\!\lower0.7ex\hbox{$3$}}}&{{\raise0.7ex\hbox{$1$} \!\mathord{\left/
 {\vphantom {1 3}}\right.\kern-\nulldelimiterspace}
\!\lower0.7ex\hbox{$3$}}}&{{\raise0.7ex\hbox{$1$} \!\mathord{\left/
 {\vphantom {1 4}}\right.\kern-\nulldelimiterspace}
\!\lower0.7ex\hbox{$4$}}}&{{\raise0.7ex\hbox{$1$} \!\mathord{\left/
 {\vphantom {1 6}}\right.\kern-\nulldelimiterspace}
\!\lower0.7ex\hbox{$6$}}} \\ 
  {{\raise0.7ex\hbox{$1$} \!\mathord{\left/
 {\vphantom {1 6}}\right.\kern-\nulldelimiterspace}
\!\lower0.7ex\hbox{$6$}}}&{{\raise0.7ex\hbox{$1$} \!\mathord{\left/
 {\vphantom {1 4}}\right.\kern-\nulldelimiterspace}
\!\lower0.7ex\hbox{$4$}}}&{{\raise0.7ex\hbox{$1$} \!\mathord{\left/
 {\vphantom {1 3}}\right.\kern-\nulldelimiterspace}
\!\lower0.7ex\hbox{$3$}}}&{{\raise0.7ex\hbox{$2$} \!\mathord{\left/
 {\vphantom {2 3}}\right.\kern-\nulldelimiterspace}
\!\lower0.7ex\hbox{$3$}}}&{{\raise0.7ex\hbox{$3$} \!\mathord{\left/
 {\vphantom {3 4}}\right.\kern-\nulldelimiterspace}
\!\lower0.7ex\hbox{$4$}}}&{{\raise0.7ex\hbox{$5$} \!\mathord{\left/
 {\vphantom {5 6}}\right.\kern-\nulldelimiterspace}
\!\lower0.7ex\hbox{$6$}}} 
\end{array}} \right)^T}$ to generate three images between each two reference frames.

%

To benchmark our model, we chose several state-of-the-art methods in the field which are EpicFlow \cite{rn159}, a state-of-the-art optical flow method, DVF \cite{rn132}, an implicit optical flow based method, Beyond MSE \cite{rn143}, a state-of-the-art pixel hallucination method, and phase-based frame interpolation \cite{rn139}, a plug-and-play and non-learning method.

\subsection{Quantitative results}
\begin{table*}[tp]
\centering
\caption{Quantitative performances on UCF-101. Learning methods are in italic. Plug-and-play methods are underlined. Higher score is better. Best scores are in boldface.}
\label{quantitative}
\begin{tabular}{|c|c|c|c|c|c|}
\hline
Method                  & SSIM          & VIF           & DLM           & NQM           & PSNR          \\ \hline
\textit{Beyond MSE}     & 0.93          & -             & -             & -             & 32.8          \\
\textit{EpicFlow-based} & 0.95          & -             & -             & -             & 34.2          \\
\textit{DVF}            & 0.96          & 0.62          & 0.93          & 23.2          & \textbf{35.8} \\ \hline
\underline{Phase-based}       & 0.88          & 0.69          & 0.96          & 26.9          & 27.9          \\
\underline{Ours}              & \textbf{0.97} & \textbf{0.71} & \textbf{0.98} & \textbf{27.4} & 33.1          \\ \hline
\end{tabular}
\end{table*}

\begin{figure*}[tp]
\centering
\includegraphics[width=\textwidth]{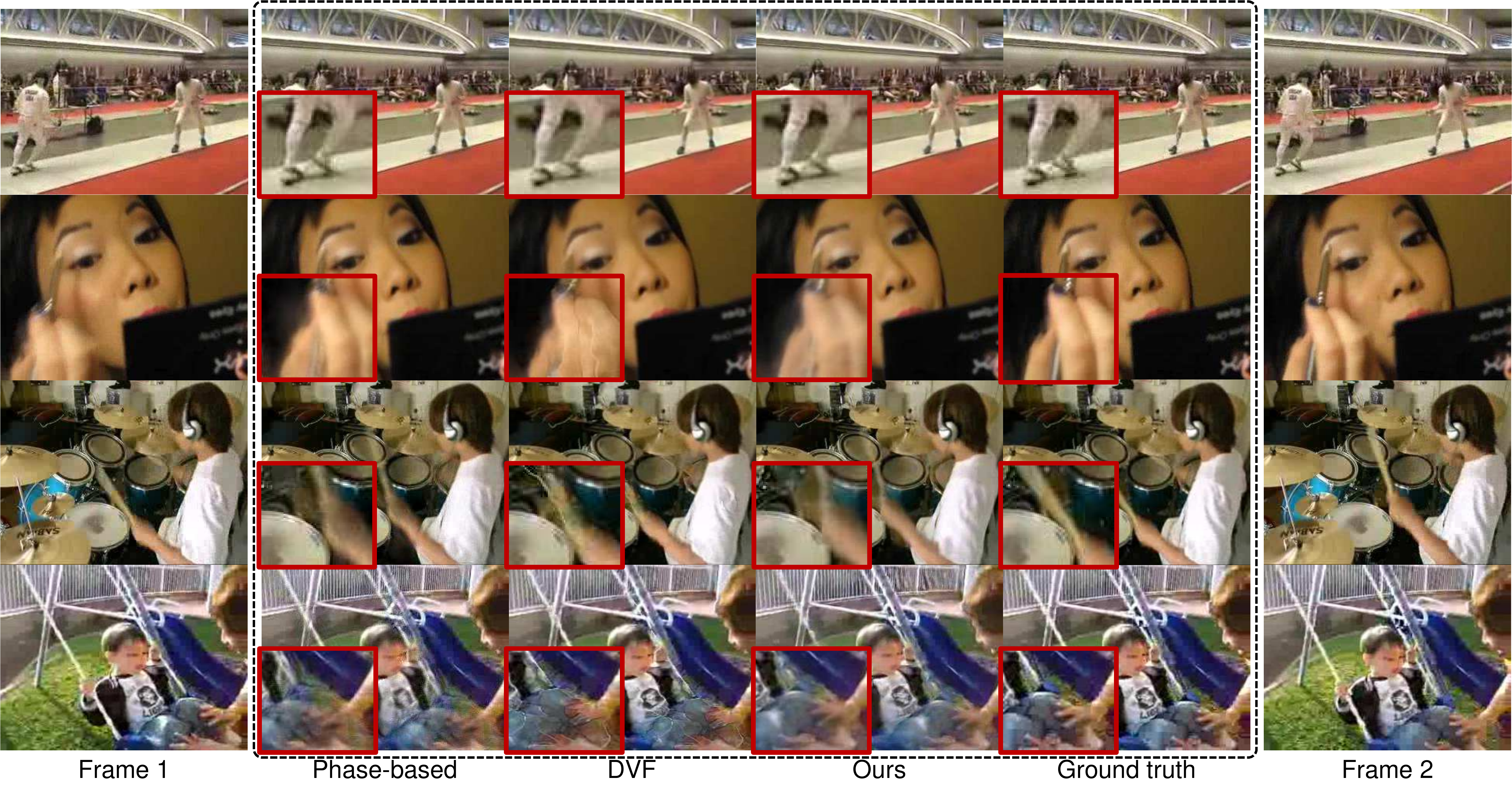}
\caption{Visual results of the benchmarking methods. Best viewed in color and zoom.}
\label{qualitative1}
\end{figure*}

Table~\ref{quantitative} shows the benchmark on UCF-101 between our method versus the chosen models. As we can see, our method outperforms all other benchmarking methods in terms of perceptual quality. We optimize the network with SSIM so it is not surprising our SSIM results are highest among all methods. That is why we employ other perceptual quality metrics for a fair benchmark with DVF. However, our synthesized images still have higher scores than DVF. The only score we lose to DVF is PSNR. However, there might exist several correct interpolated frames between any two frames so a pixel-to-pixel difference metric like PSNR cannot judge the correctness of the interpolated motions. In addition, it is well-known that PSNR often does not correlate well with humans’ opinions, hence it usually yields poor performances in quality assessment studies \cite{rn169}. On the other hand, SSIM, VIF, DLM and NQM are famous for their correlations with the quality that humans perceive. Having high performances on these metrics is a strong indicator that our generated images are likely to be favored by viewers. 

\subsection{Qualitative results} 


Figure~\ref{qualitative1} demonstrates the interpolated frames of DVF, the phase-based method and ours, along with the corresponding ground truths. In the first row, when the movement (of the heel of the left person) is medium, all methods perform well but the phase-based still leaves visible artifacts. In the second row, DeepLLE and the phase-based produce similar results, but DVF generates artifacts even though the motion of the hand is very simple. This is perhaps due to the motion blur in the first frame, and this reveals that DVF may not succeed when there is some abrupt intensity change between the two inputs. In the third row, when the motions are large, all methods struggle to interpolate the in-between frames. While the phase-based's results and ours are still reasonable, the DVF's result is distorted because of the incorrect image warping. This sort of distortion is typical in optical flow based methods. In the last row, when motions are too large, all methods fail. Our method, however, still manages to generate an image with heavy blur in the motion regions while DVF confusingly warps the image and the result is unrecognizable. It can be seen that the images generated by our method are of equal quality compared with even state-of-the-art learning-based methods. Also, taking a closer look, we can notice that our method actually reduces artifacts and distortions in the original images. The same phenomenon has been observed and discussed in \cite{rn145}, a study concurrent to ours. We conclude that our method is more preferred than existing methods in many situations since the method does not need pre-training on big data and works like conventional methods but at the same time possesses the computing power of deep learning which greatly improves the quality of the synthesized images over conventional models and makes it on par with state-of-the-art deep learning based methods.

\begin{figure*}[tp]
\centering
\includegraphics[width=\textwidth]{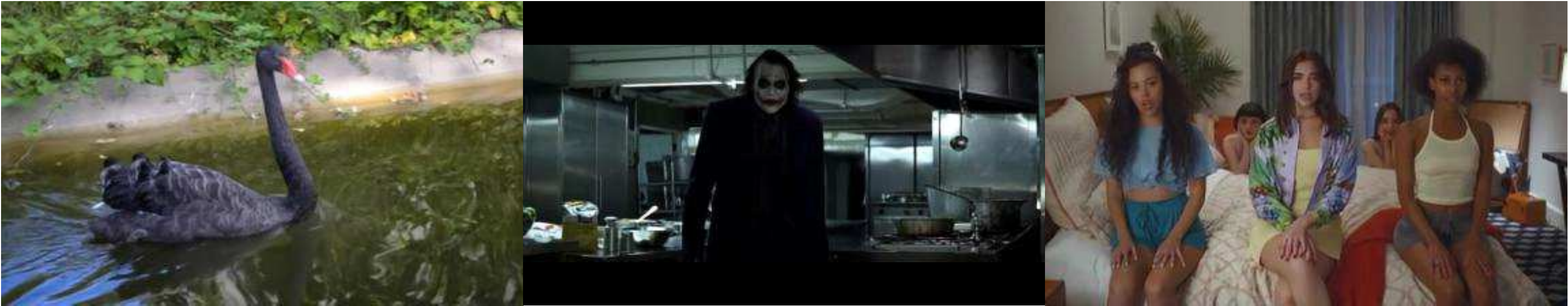}
\caption{Examples of our interpolated frames in DAVIS and real-life videos. Best viewed in color and zoom.}
\label{qualitative2}
\end{figure*}

Figure~\ref{qualitative2} shows some of our interpolated frames from several DAVIS and real-life videos. We processed several segments of the videos and ran them at $12$ $fps$ for slow-motion effect. We highly encourage readers to check our webpage\textsuperscript{\ref{footnote supp}} for more results since it is impossible to evaluate temporal coherence on still paper.

\subsection{Extension to extrapolation}

\begin{figure}[tp]
\centering
\includegraphics[width=\textwidth]{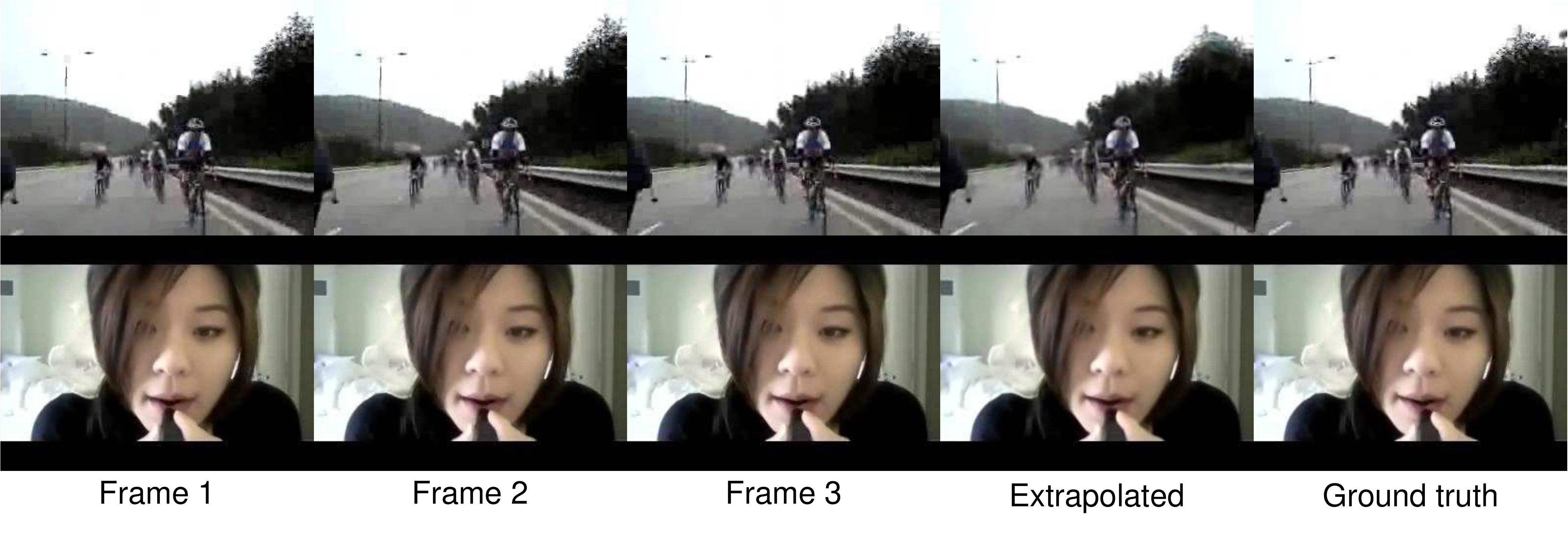}
\caption{Video frames extrapolated by our method. Best viewed in color and zoom.}
\label{ext}
\end{figure}

Our framework can be easily extended to video frame extrapolation. The setup of the experiment is described in our supplementary material. Figure~\ref{ext} displays the future frames synthesized by our method. Our observation is that in many sequences the generated images do not show clear motions. Even if the motions in the reference frames are large, the generated motions are still limited. This suggests that extending our framework to extrapolation is a non-trivial task which we will take on in our future work.

\subsection{Ablation study}

\begin{table}[tp]
\centering
\caption{Quantitative ablation study on different components of our framework.}
\label{ablation}
\resizebox{\textwidth}{!}{\begin{tabular}{|c|c|c|c|c|c|c|c|c|}
\hline
Ablation & \begin{tabular}[c]{@{}c@{}}Ours\\ (4 ref. frames)\end{tabular} & \begin{tabular}[c]{@{}c@{}}Ours\\ (w/o LLE)\end{tabular} & \begin{tabular}[c]{@{}c@{}}Ours\\ (w/o dropout)\end{tabular} & \begin{tabular}[c]{@{}c@{}}Ours\\ (w/o SSIM)\end{tabular} & \begin{tabular}[c]{@{}c@{}}Ours\\ (ReLU)\end{tabular} & \begin{tabular}[c]{@{}c@{}}Ours\\ (ELU)\end{tabular} & \begin{tabular}[c]{@{}c@{}}Ours\\ (SELU)\end{tabular} & Ours          \\ \hline
SSIM     & 0.96                                                           & 0.94                                                     & 0.95                                                         & 0.95                                                      & 0.96                                                  & 0.96                                                 & 0.95                                                  & \textbf{0.98} \\
VIF      & 0.70                                                           & 0.59                                                     & 0.60                                                         & 0.64                                                      & 0.61                                                  & 0.69                                                 & 0.64                                                  & \textbf{0.71} \\
DLM      & 0.96                                                           & 0.92                                                     & 0.93                                                         & 0.95                                                      & \textbf{0.97}                                         & 0.96                                                 & 0.96                                                  & \textbf{0.97} \\
NQM      & 28.0                                                           & 22.4                                                     & 24.0                                                           & 27.3                                                      & \textbf{30.9}                                         & 25.7                                                 & 27.4                                                  & 27.3          \\
PSNR     & 30.9                                                           & 27.2                                                     & 28.0                                                         & 32.5                                                      & 31.5                                                  & 29.6                                                 & 30.7                                                  & \textbf{33.4} \\ \hline
\end{tabular}}
\end{table}

\begin{figure*}[tp]
\centering
\includegraphics[width=\textwidth]{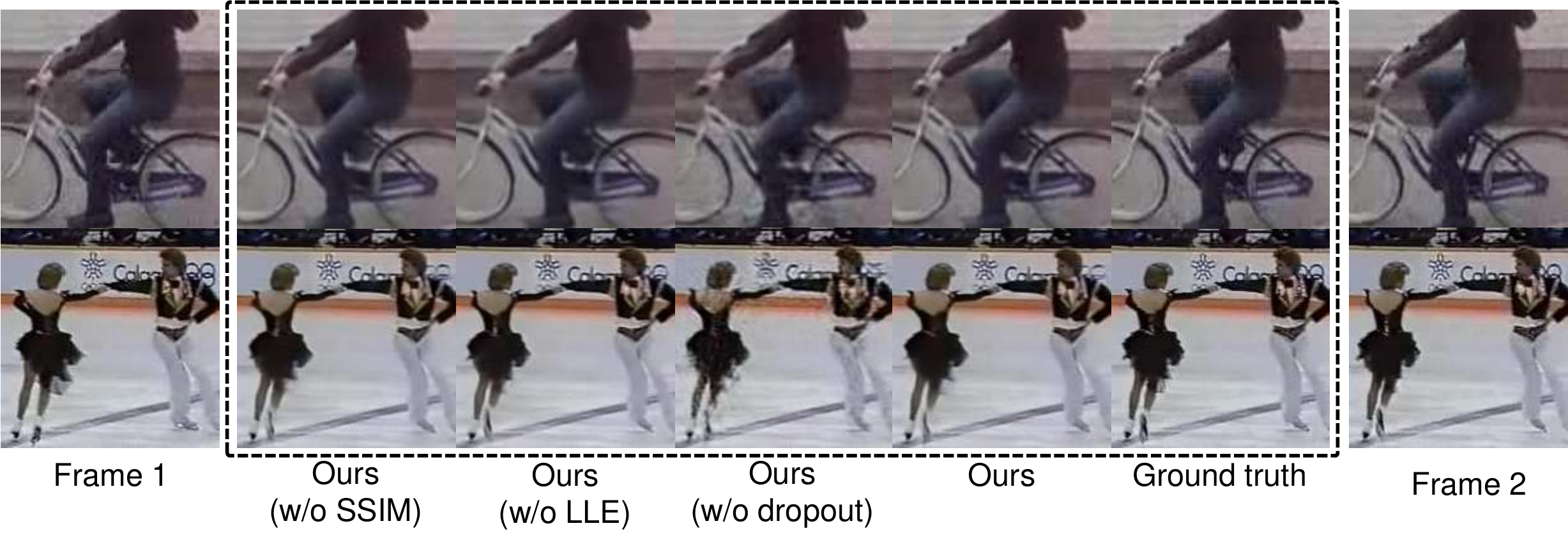}
\caption{Qualitative ablation study on different components of our framework.}
\label{ablation1}
\end{figure*}

We verify some of our choices for the proposed framework. To perform the ablation study, we randomly selected $20$ videos from the UCF-101 dataset. All settings except for the ablated components were set to our defaults.

\textbf{Locally linear embedding module.} \label{ablation LLE}
To study the effect of the locally linear embedding module, we train an auto-encoding CNN by a common training scheme. Concretely, we remove the LLE module and use all the reference frames as input. The network architecture is the same as in Section~\ref{learning}. Table~\ref{ablation} and Figure~\ref{ablation1} show the quantitative and qualitative performances when the LLE module is removed in training. In general, the quality of the generated images is blurrier than our full framework. In some sequences, the synthesized images do not contain any motion. They look almost the same as either the two nodes. The numerical results consensually suggest that without LLE, the quality is much worse. We can see that using only two reference frames ($N=1$) is in fact equivalent to this case. Therefore, our method can work with only two input frames by discarding the LLE module. However, this is highly discouraged. Without the module, the network is not taught how to decode the points on the line segment passing through the two nodes. By explicitly embedding the linearity on the latent manifold, we guide the network to encode the nodes so that the decoding of the synthesized latent codes can be performed properly. 

Nevertheless, an important observation from this benchmark is that the deep network captures a great deal of the underlying manifolds of video sequences since it still can manage to decode an interpolated latent variable even though it is not optimized to do so. In \cite{rn145}, the authors discovered that deep networks can capture the structure of an image, and through that they can blindly restore or super-resolve an image. Here, we have found that in addition to image structure, they can capture video structure as well. This analysis can be greatly beneficial to the advancement of the deep unsupervised learning.

\begin{wrapfigure}{R}{0.5\textwidth}
\centering
\includegraphics[width=0.5\textwidth]{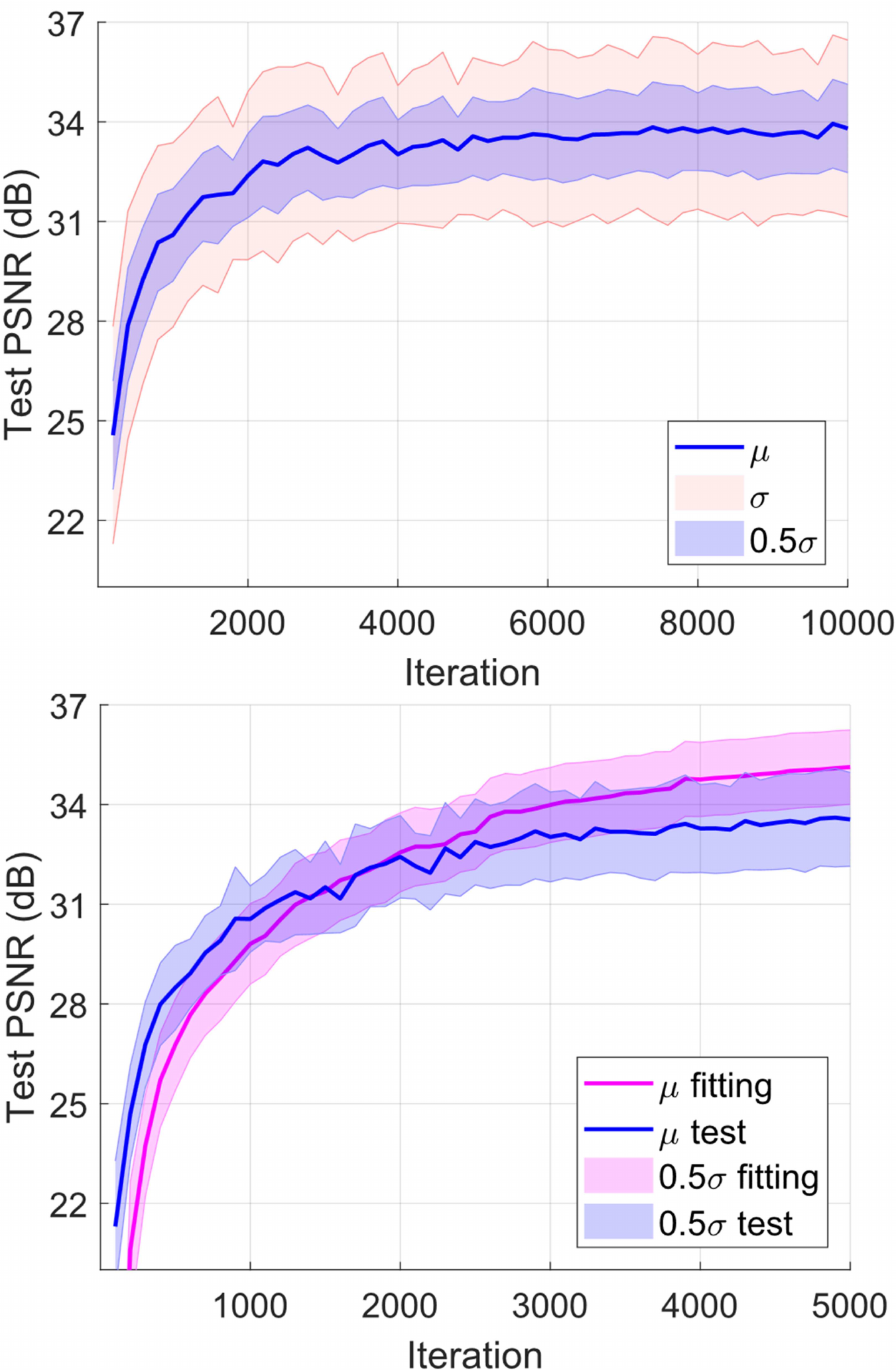}
\caption{Effect of number of iterations (upper) and learning rate adjustment (lower) on the test PSNR scores. Best viewed in color and zoom.}
\label{plot}
\end{wrapfigure}


\textbf{Number of reference frames.}
We conducted an ablation study on the number of reference frames. Table~\ref{ablation} demonstrates our numerical results when the number of reference images is $4$. As can be seen, the results are much worse than the three reference frame case. However, readers might take this benchmark with a grain of salt since in this case, the two nodes are 5 frames apart, which may contain too large a motion. Nevertheless, there is a trade-off between the number of references and the linearity assumption. In practice, the manifold should be highly non-linear, and the latent codes of the references may not even be co-planar, much less on the same line. That is why, keeping the number of references small might result in a better performance. Further, using three frames is more computationally economical, so we use three reference frames as a default setting.

\textbf{Effect of number of iterations.} Figure~\ref{plot} displays the test PSNR values when the number of iterations is $10000$. In our experiments, the general tendency is that the longer the optimization, the better the performance. Different from existing methods, ours blindly optimizes the cost function and also, there is no data distribution to generalize. In that sense, our method is closer to pure optimization than learning, which requires a minimization of an empirical risk. Thus, an optimization of at least $5000$ iterations is recommended for a satisfying result.

\textbf{Effect of learning rate adjustment.} We experimented with a manual learning rate adjustment scheme in which we halved the learning rate at iterations $2500$ and $3750$. The results of such a scheme are shown in Figure~\ref{plot}. Although the scheme does not help improve the overall result, it does stabilize the optimization when the network is near its saturation point. Unlike the case without any learning rate adjustment in which the curves fluctuate strongly, this scheme guarantees a good solution no matter when the optimization is terminated. Therefore, it is advised to turn this scheme on although it is only optional.

\textbf{Effect of dropout.}
Figure~\ref{ablation1} shows the visual results of our method with and without dropout. As can be seen from the figure, the images generated without dropout have noticeable artifacts in the interpolated regions while with dropout do not show any sort of artifacts. Moreover, the network with dropout recovers high-frequency components better, which makes the images more realistic and less blurry. Table~\ref{ablation} shows the numerical results of our ablation study on dropout. It is clear that the performance of the network without dropout is nowhere near the true performance. We conclude that dropout can significantly improve the quality of the synthesized images.

\textbf{Choice of activation.}
As can be seen from Table~\ref{ablation}, our method using LReLU surpasses all other configurations of activation functions. An explanation is that ReLU, SELU and ELU saturate when the argument is below some threshold. This saturation makes the network robust against small perturbations in input \cite{rn156}, which is quite unwanted in our method since the two nodes' latent variables and the interpolated codes are very close to each other. This experiment implies that although deep networks is a powerful computing tool, one may need to design their architectures based on thorough analyses of the problems that they are applied to.

\textbf{Effect of SSIM cost.}
From Table~\ref{ablation}, we can see that when excluding SSIM from the cost function, SSIM and PSNR slightly drop. The visual results can be seen in Figure~\ref{ablation}. At first, we do not find much difference between the images optimized with and without SSIM but upon closer examination, we can see several artifacts in the images. Since SSIM is a structural similarity metric, including it in the cost function may prevent the network from generating structural artifacts, which makes the synthesized images better-looking. 

\textbf{Limitations.}
We observe that our method completely fails when the input images are mostly black such as some UCF-101 \textit{PlayingPiano} and \textit{PlayingFlute} sequences. Also, due to the nature of the method, it cannot be used in an online scenario since the network has to perform optimization for each input sequence.

\section{Conclusion}
In this paper, we present a new method for in-between frame generation. Our method is constructed by embedding an explicit linearity assumption to the latent representations of consecutive frames obtained from an auto-encoding CNN. Our model is an off-the-shelf method that can be applied instantly to a video like conventional methods while possessing deep-learning-level performances. Our model can synthesize an arbitrary number of images between any number of frames simultaneously. Different benchmarks reveal that our synthesized images are comparable with state-of-the-art performances in the field. Moreover, our work exposes that deep CNNs capture a great deal of video statistics, which may be greatly beneficial to the study of deep unsupervised learning.

\bibliographystyle{splncs}

\end{document}